\documentclass[12pt]{article}
\usepackage[utf8]{inputenc}
\usepackage{amsmath, amssymb}
\usepackage{graphicx}
\usepackage{geometry}
\usepackage{titlesec}
\usepackage{enumitem}
\usepackage{setspace}
\usepackage{fancyhdr}
\usepackage{hyperref}
\usepackage{float}
\usepackage{makecell} 
\usepackage{subcaption}

\titleformat{\section}{\large\bfseries}{\thesection.}{1em}{}
\titleformat{\subsection}{\normalsize\bfseries}{\thesubsection.}{1em}{}

\title{Investigating Optical Flow Computation: From Local Methods to a Multiresolution Horn-Schunck Implementation with Bilinear Interpolation}
\author{Haytham Ziani \\
Al Akhawayn University in Ifrane \\
h.ziani@aui.ma}
\date{\today}

\begin{document}

\maketitle

\begin{abstract}
This paper presents an applied analysis of local and global methods, with a focus on the Horn-Schunck algorithm for optical flow computation. We explore the theoretical and practical aspects of  local approaches, such as the Lucas-Kanade method, and global techniques such as Horn-Schunck. Additionally, we implement a multiresolution version of the Horn-Schunck algorithm, using bilinear interpolation and prolongation to improve accuracy and convergence. The study investigates the effectiveness of these combined strategies in estimating motion between frames, particularly under varying image conditions.
\end{abstract}

\noindent\textbf{Keywords:} Optical flow, Bilinear Optimization, Horn-Schunck Algorithm, Multiresolution Methods

\section{Introduction}
Optical flow estimation is a fundamental task in computer vision concerned with determining the apparent motion of brightness patterns between consecutive frames in a video sequence. The goal is to compute a dense motion field that describes how each pixel moves over time. This problem is critical in numerous applications including motion detection, object tracking, video compression, and autonomous navigation.

From a mathematical standpoint, most optical flow estimation techniques are built upon the \textit{brightness constancy assumption}, which states that the intensity of a moving point in the scene remains constant over time. Given an image sequence \( I(x, y, t) \), the brightness constancy can be expressed as:
\begin{equation}
I(x, y, t) = I(x + u, y + v, t + 1),
\end{equation}
where \( (u, v) \) is the optical flow vector at location \( (x, y) \). By applying a first-order Taylor expansion and assuming small displacements, we obtain the classical Optical Flow Constraint Equation (OFCE):
\begin{equation}
I_x u + I_y v + I_t = 0,
\end{equation}
where \( I_x, I_y \) and \( I_t \) denote the partial derivatives of the image intensity with respect to spatial and temporal coordinates, respectively. This equation provides a single constraint for two unknowns per pixel, making the system underdetermined — known as the \textit{aperture problem}.

To address this, various methods have been developed. In this paper, we analyze the effectiveness of local and global methods, focusing on the effectiveness of a multiresolution Horn-Schunck framework with bilinear interpolation for prolongation between pyramid levels. This approach improves accuracy in large-displacement scenarios and enhances computational efficiency through coarse-to-fine estimation strategies.

\section{Sparse Methods}

Sparse optical flow methods, also known as local methods, estimate motion by analyzing small patches or neighborhoods around selected feature points in the image. Rather than attempting to compute flow vectors for every pixel, these approaches focus on points where reliable motion estimation is possible, typically at locations with sufficient texture, such as corners or high-contrast regions. These methods rely on the assumption that within a small neighborhood, the motion (optical flow) is constant, and thus can be inferred from the local variations in intensity over time. 

One of the most widely used local methods is the Lucas-Kanade algorithm, which operates by applying the optical flow constraint equation (OFCE) over a window \( \Omega_p \) centered at each pixel \( p \). The OFCE is given by:
\begin{equation}
I_x u + I_y v + I_t = 0,
\end{equation}
where \( I_x \) and \( I_y \) are the spatial image gradients in the \( x \) and \( y \) directions respectively, \( I_t \) is the temporal gradient (intensity change between frames), and \( u \), \( v \) represent the components of the flow vector. Since a single equation is insufficient to determine two unknowns, the method collects this equation over all pixels within the neighborhood \( \Omega_p \), resulting in an overdetermined system of the form:
\begin{equation}
A w = b,
\end{equation}
where \( A \in \mathbb{R}^{n \times 2} \) contains the stacked gradients \( [I_x, I_y] \) of the \( n \) pixels in the neighborhood, \( w = [u, v]^T \) is the flow vector to be estimated, and \( b \in \mathbb{R}^{n} \) contains the negative temporal gradients \( -I_t \). The flow is then estimated using the least-squares solution:
\begin{equation}
w(p) = (A^T A)^{-1} A^T b.
\end{equation}

This solution is valid and stable when the matrix \( A^T A \) is invertible, which requires that the local image patch contains sufficient intensity variation in multiple directions—commonly referred to as having a strong structure tensor. When this condition is not met, the flow estimation may become unreliable. To improve robustness, the Lucas-Kanade method is often combined with feature selection algorithms (e.g., Shi-Tomasi corner detection) and can be extended to a multiscale (pyramidal) version to handle larger displacements. Although sparse methods are generally faster and less memory-intensive than dense approaches, they only provide motion estimates at selected points and may miss important flow information in textureless regions.

\section{Dense Methods}

\subsection{Horn-Schunck Method}

Global methods, on the contrary, incorporate a regularization term throughout the image domain to enforce spatial coherence. The Horn-Schunck method is one of the earliest and most influential global approaches to optical flow estimation. It addresses the underdetermined nature of the Optical Flow Constraint Equation (OFCE) by introducing a global smoothness prior that enforces the assumption that neighboring pixels in the flow field should have similar motion vectors.

\subsubsection*{Mathematical Formulation}

The method minimizes the following energy functional over the entire image domain \( \Omega \):
\begin{equation}
E(u, v) = \int_\Omega (I_x u + I_y v + I_t)^2 + \alpha \left( |\nabla u|^2 + |\nabla v|^2 \right) \, dxdy,
\end{equation}
where:
\begin{itemize}
    \item \( (u, v) \) is the optical flow field,
    \item \( I_x, I_y, I_t \) are the partial derivatives of the image intensity with respect to \( x, y \), and \( t \),
    \item \( \nabla u \) and \( \nabla v \) are the gradients of the flow fields, and
    \item \( \alpha \) is a regularization parameter that controls the trade-off between data fidelity and smoothness.
\end{itemize}

The first term, known as the data term, enforces the brightness constancy constraint. The second term is a smoothness constraint, which penalizes large variations in the flow field, encouraging a globally smooth flow. This is especially helpful in homogeneous regions with little texture where the data term alone would be ambiguous.

To minimize this energy, Horn and Schunck derived the Euler-Lagrange equations:
\begin{equation}
I_x (I_x u + I_y v + I_t) - \alpha \nabla^2 u = 0,
\end{equation}
\begin{equation}
I_y (I_x u + I_y v + I_t) - \alpha \nabla^2 v = 0,
\end{equation}
where \( \nabla^2 \) denotes the Laplacian operator. These equations are typically solved iteratively using successive over-relaxation or other numerical schemes, updating each flow component based on local averages and the image derivatives.

In this implementation of the Horn-Schunck optical flow algorithm, numerical computations were performed using an iterative Gauss-Seidel method. The algorithm relies on the brightness constancy constraint, enforced via image gradients computed from the first frame. The spatial derivatives \( I_x \) and \( I_y \) were obtained using Sobel operators with a kernel size of \( 3 \times 3 \), while the temporal derivative \( I_t \) was calculated as the pixel-wise difference between the second and first frames: \( I_t = I_2 - I_1 \). The initial velocity fields \( u \) and \( v \) were set to zero and updated iteratively. 

To approximate the Laplacian regularization term, Gaussian smoothing was applied using a \( 5 \times 5 \) kernel, replacing the standard discrete Laplacian stencil. This smoothing process produced local averages \( \bar{u} \) and \( \bar{v} \), which were used in the update equations. At each iteration, the flow fields were updated according to:

\[
u = \bar{u} - \frac{I_x (I_x \bar{u} + I_y \bar{v} + I_t)}{\alpha^2 + I_x^2 + I_y^2 + \epsilon}, \quad
v = \bar{v} - \frac{I_y (I_x \bar{u} + I_y \bar{v} + I_t)}{\alpha^2 + I_x^2 + I_y^2 + \epsilon}
\]

where \( \alpha \) is the regularization parameter and \( \epsilon \) is a small constant to prevent division by zero. The algorithm iterates until convergence, defined as the \( L^2 \)-norm of the flow change falling below a threshold \( \epsilon = 10^{-5} \), or until the maximum number of iterations (5000) is reached.

Although the Horn-Schunck method provides dense flow estimates and handles large smooth regions well, it struggles at motion boundaries due to the quadratic nature of the regularization. This smoothness prior tends to blur discontinuities in the motion field, leading to inaccurate flow estimates near object edges and occlusion boundaries.

Another significant limitation arises in the presence of large displacements. The derivation of the Horn-Schunck method is based on the assumption of small motion between consecutive frames, which justifies the use of the first-order Taylor expansion of the image brightness constancy assumption. However, when the true motion between frames is large, this linear approximation breaks down. Consequently, the algorithm may converge to a local minimum that poorly represents the actual flow, resulting in degraded accuracy.

Moreover, image gradients used in the data term \( (I_x u + I_y v + I_t)^2 \) are local in nature and may not capture broader contextual cues necessary for estimating large motions. The algorithm also performs fixed-size averaging in the smoothness term, which may not be sufficient to bridge distant correspondences required for accurate large-displacement flow estimation.

This challenge has motivated the development of hierarchical or multiresolution approaches, which will be discussed in a later section. These methods progressively estimate the flow from coarse to fine image resolutions, enabling the capture of large displacements in coarse levels and refining the details in finer levels.

\subsection{Multiresolution Horn-Schunck}

To improve the original Horn-Schunck algorithm’s ability to estimate large motions, a multiresolution or coarse-to-fine strategy is commonly adopted. This approach constructs a pyramid of downsampled images, where optical flow is first computed at the coarsest level and progressively refined at higher resolutions. The flow field estimated at one level is upsampled to the next finer level and used to initialize further iterations of the Horn-Schunck algorithm.

This hierarchical framework significantly enhances robustness and convergence by avoiding poor local minima and providing a strong prior at each level. Additionally, it maintains computational efficiency since early estimates are obtained on smaller image representations.

\subsubsection*{Pyramid Construction and Prolongation with Bilinear Interpolation}

Given a pair of input images \( I_t \) and \( I_{t+1} \), we construct a Gaussian pyramid \( \{I^0, I^1, \dots, I^L\} \), where \( I^0 \) is the original image and \( I^L \) is the coarsest version. The pyramid is formed by iteratively applying Gaussian smoothing and downsampling, typically by a factor of 2 at each level.

Once the optical flow \( (u^l, v^l) \) is computed at a coarse level \( l \), it needs to be upsampled to level \( l-1 \) to initialize the flow estimate there. This process is called \textit{prolongation}, and it is done using \textbf{bilinear interpolation}.

\paragraph{Mathematical Formulation of Bilinear Interpolation}

Let \( f(i, j) \) denote a discrete 2D function (such as a flow field or image), and suppose we wish to estimate its value at a continuous coordinate \( (x, y) \), where \( x \in [i, i+1) \) and \( y \in [j, j+1) \). The bilinear interpolation of \( f \) at \( (x, y) \) is given by:

\begin{equation}
f(x, y) \approx (1 - a)(1 - b) f(i, j) + a(1 - b) f(i+1, j) + (1 - a)b f(i, j+1) + ab f(i+1, j+1),
\end{equation}

where \( a = x - i \) and \( b = y - j \) are the fractional parts. This method performs linear interpolation first in one direction and then in the other. It assumes continuity and linearity between pixel values, producing smooth and reasonably accurate estimates at non-integer positions.

\paragraph{Application in Flow Prolongation}

In the context of multiresolution optical flow, bilinear interpolation is used to upsample the flow vectors \( (u^l, v^l) \) from level \( l \) to the higher-resolution level \( l-1 \). Since flow fields typically have floating-point values and non-integer displacements, bilinear interpolation is well suited to maintain smoothness and spatial coherence across resolutions.

\paragraph{Application in Image Warping}

The upsampled flow is also used to warp the second image \( I^{l-1}_{t+1} \) toward the first image \( I^{l-1}_t \), aligning them under the current flow estimate. Warping is implemented via inverse mapping: for each pixel \( (x, y) \), we evaluate the intensity at location \( (x + u(x, y), y + v(x, y)) \) in the second image. Since these coordinates are usually non-integer, bilinear interpolation is applied to obtain the pixel value:

\begin{equation}
I_{t+1}^{\text{warped}}(x, y) = I_{t+1}^{l-1}(x + u(x, y), y + v(x, y)).
\end{equation}

In our implementation, OpenCV’s \texttt{cv2.remap()} function with \texttt{INTER\_LINEAR} mode performs this interpolation. This ensures subpixel accuracy during warping and helps preserve the brightness constancy assumption, which is essential for the reliability of the Horn-Schunck model.

\subsubsection*{Algorithm Summary}

\begin{enumerate}[label=(\alph*)]
    \item Build Gaussian pyramids of the input image pair.
    \item Initialize flow \( (u^L, v^L) = (0, 0) \) at the coarsest level.
    \item For each level \( l = L, L-1, \dots, 0 \):
    \begin{itemize}
        \item Upsample flow \( (u^l, v^l) \) to level \( l-1 \) using bilinear interpolation.
        \item Warp the second image at level \( l-1 \) using the upsampled flow.
        \item Compute spatial and temporal gradients.
        \item Refine flow via Horn-Schunck iterations.
    \end{itemize}
    \item Return the refined flow \( (u^0, v^0) \) at full resolution.
\end{enumerate}

The incorporation of bilinear interpolation in both flow prolongation and image warping is essential for preserving accuracy and smoothness across pyramid levels. Its low computational cost and effective handling of subpixel precision make it a fundamental component of modern optical flow pipelines.

\section{Experimental Results on the Sintel Dataset}

The Sintel dataset is a synthetic optical flow benchmark derived from the open-source animated short film \textit{Sintel} by the Blender Foundation. It provides realistic and complex scenes featuring large displacements, motion blur, atmospheric effects, and illumination changes. Each frame comes with ground truth optical flow, occlusion masks, and disparity maps, making it widely used for training and evaluating dense optical flow algorithms. The dataset includes both a ``clean'' pass (with minimal post-processing) and a ``final'' pass (with full cinematic rendering).

To evaluate the performance of the Horn-Schunck algorithm, we applied it to several ``final'' frames from different scenes of the MPI Sintel dataset, a widely used benchmark for optical flow evaluation that includes complex scenes with non-rigid motion, occlusions, and illumination effects. The algorithm was implemented in Python and executed using Google Colab, which provided a convenient and accessible cloud-based environment for development and testing.

We report the \textbf{Average Angular Error (AAE)} and the \textbf{End-Point Error (EPE)} for selected frames in the table.  
The AAE measures the average angular difference between the estimated optical flow vector $(u, v)$ and the ground truth vector $(u_{gt}, v_{gt})$ for each pixel. It is computed as:
\[
\text{AAE} = \frac{1}{N} \sum_{i=1}^N \cos^{-1}\left( \frac{u_i u_{gt,i} + v_i v_{gt,i}}{\sqrt{u_i^2 + v_i^2 + \varepsilon} \cdot \sqrt{u_{gt,i}^2 + v_{gt,i}^2 + \varepsilon} + \varepsilon} \right)
\]
where $\varepsilon$ is a small constant to prevent division by zero and numerical instability.

The EPE quantifies the average Euclidean distance between estimated and ground truth flow vectors across all pixels:
\[
\text{EPE} = \frac{1}{N} \sum_{i=1}^N \sqrt{(u_i - u_{gt,i})^2 + (v_i - v_{gt,i})^2}
\]
Both metrics provide complementary insights into the accuracy of optical flow estimation---AAE emphasizes directional accuracy, while EPE measures overall vector magnitude error.

\begin{table}[H]
\centering
\caption{Comparison of Horn-Schunck and Multiresolution Horn-Schunck on Sintel scenes. Each cell shows \textbf{AAE (degrees)} on the first line and \textbf{EPE (pixels)} on the second.}
\vspace{0.5cm}
\label{tab:hs-vs-mrhs-stacked}
\begin{tabular}{|c|c|c|c|}
\hline
\textbf{Scene} & \textbf{HS} & \textbf{MR-HS} & \textbf{Pyramid Levels} \\
\hline
\textit{Alley\_1} & \makecell{12.46$^\circ$\\2.62 px} & \makecell{6.61$^\circ$\\1.81 px} & 4 \\
\hline
\textit{Bamboo\_2} & \makecell{10.83$^\circ$\\1.68 px} & \makecell{8.81$^\circ$\\1.17 px} & 4 \\
\hline
\textit{Market\_2} & \makecell{19.08$^\circ$\\0.47 px} & \makecell{15.31$^\circ$\\0.41 px} & 3 \\
\hline
\textit{Mountain\_1} & \makecell{17.13$^\circ$\\3.90 px} & \makecell{15.28$^\circ$\\2.78 px} & 4 \\
\hline
\textbf{Average} & \makecell{\textbf{14.88$^\circ$}\\\textbf{2.17 px}} & \makecell{\textbf{11.50$^\circ$}\\\textbf{1.54 px}} & -- \\
\hline
\end{tabular}
\end{table}

\begin{figure}[h]
    \centering
    \captionsetup[subfigure]{labelformat=empty}
    \begin{subfigure}[t]{0.24\textwidth}
        \includegraphics[width=\linewidth]{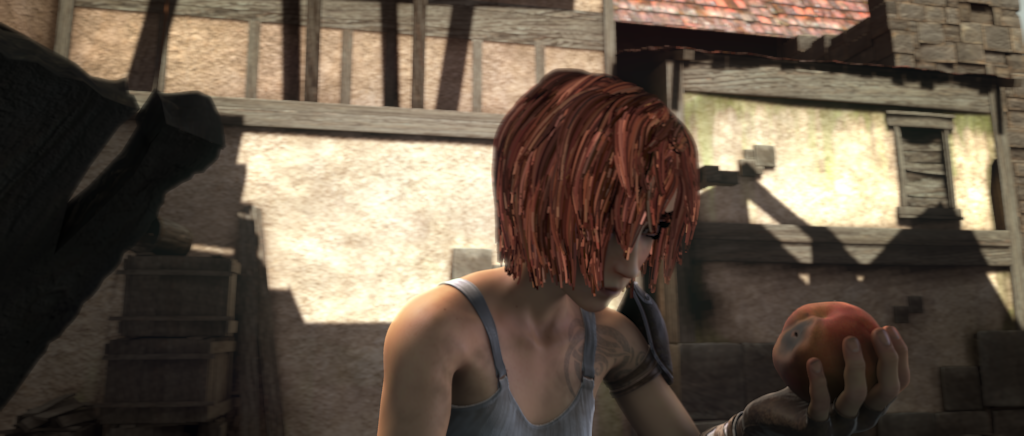}
        \caption{a) Input Frame}
    \end{subfigure}
    \hfill
    \begin{subfigure}[t]{0.24\textwidth}
        \includegraphics[width=\linewidth]{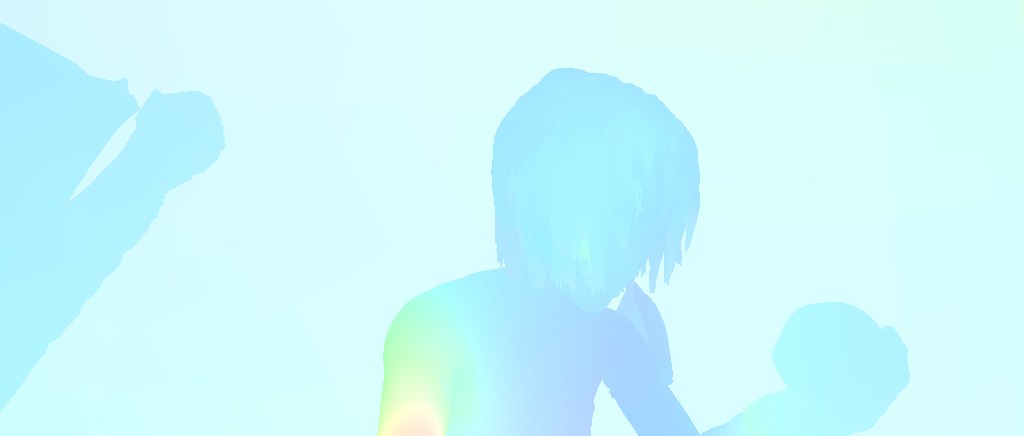}
        \caption{Ground Truth}
    \end{subfigure}
    \hfill
    \begin{subfigure}[t]{0.24\textwidth}
        \includegraphics[width=\linewidth]{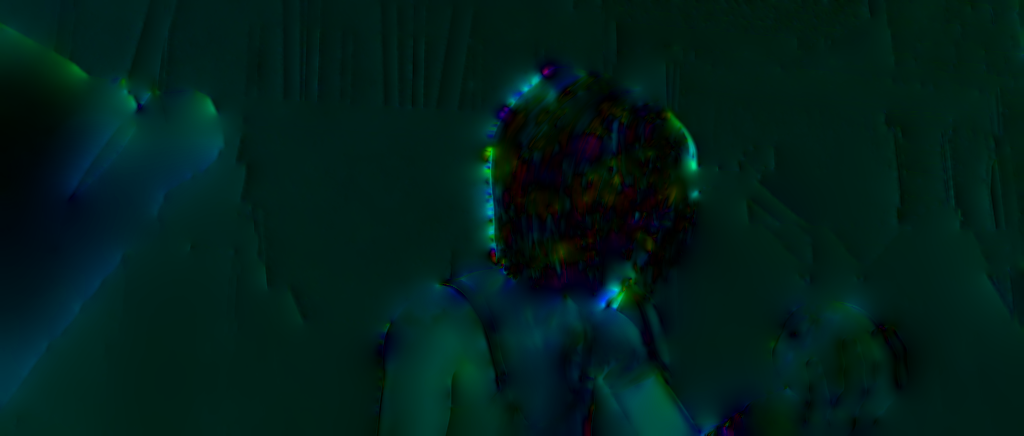}
        \caption{HS Flow}
    \end{subfigure}
    \hfill
    \begin{subfigure}[t]{0.24\textwidth}
        \includegraphics[width=\linewidth]{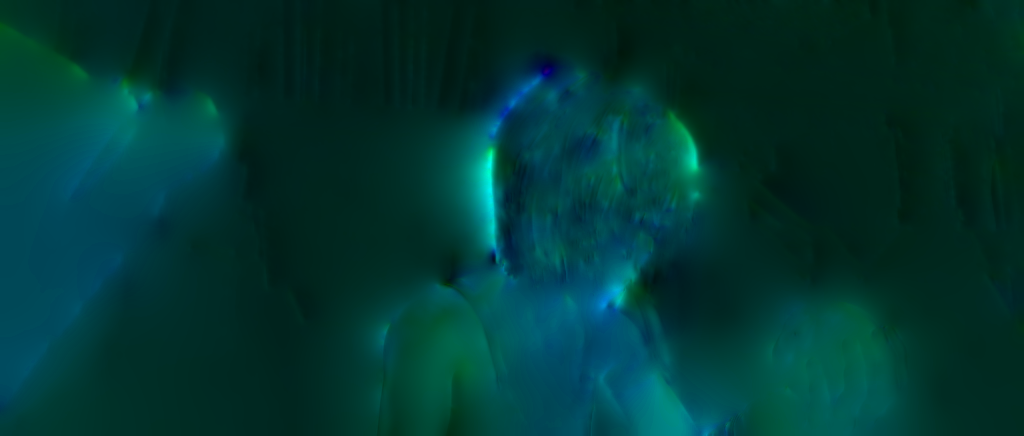}
        \caption{MR-HS Flow}
    \end{subfigure}
    \vspace{0.5em}
    \begin{subfigure}[t]{0.24\textwidth}
        \includegraphics[width=\linewidth]{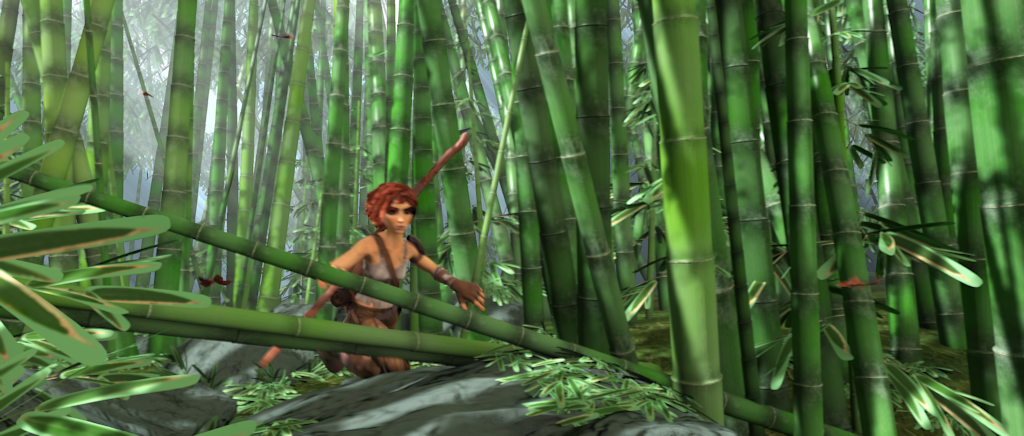}
        \caption{b) Input Frame}
    \end{subfigure}
    \hfill
    \begin{subfigure}[t]{0.24\textwidth}
        \includegraphics[width=\linewidth]{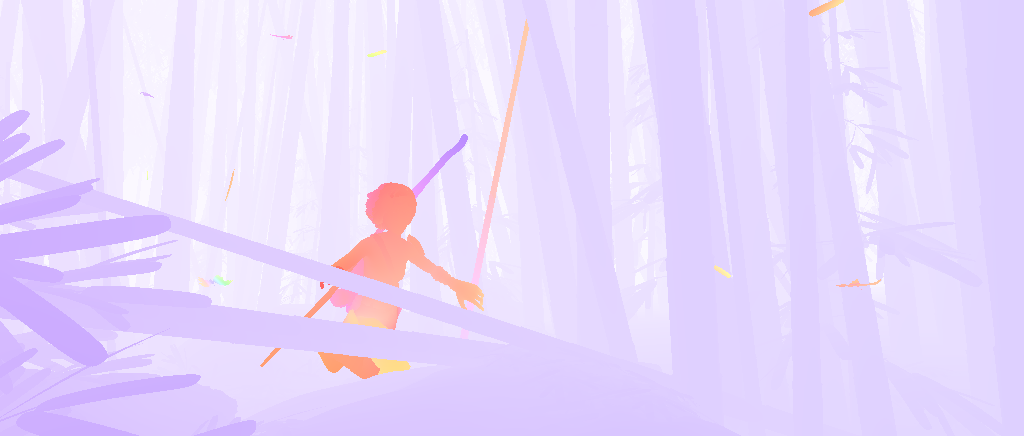}
        \caption{Ground Truth}
    \end{subfigure}
    \hfill
    \begin{subfigure}[t]{0.24\textwidth}
        \includegraphics[width=\linewidth]{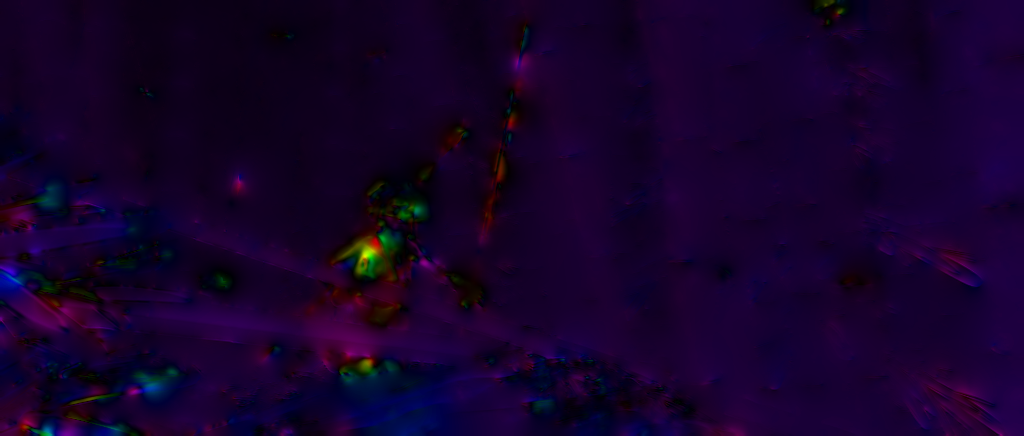}
        \caption{HS Flow}
    \end{subfigure}
    \hfill
    \begin{subfigure}[t]{0.24\textwidth}
        \includegraphics[width=\linewidth]{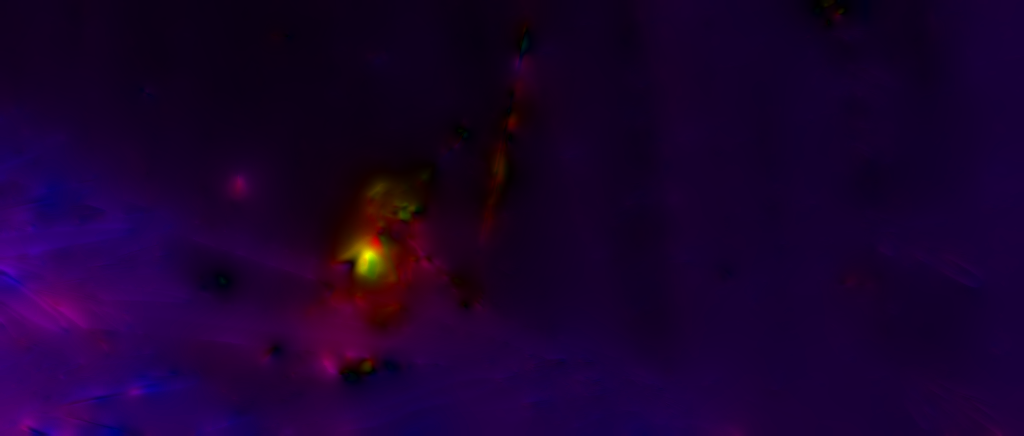}
        \caption{MR-HS Flow}
    \end{subfigure}
    \vspace{0.5em}
    \begin{subfigure}[t]{0.24\textwidth}
        \includegraphics[width=\linewidth]{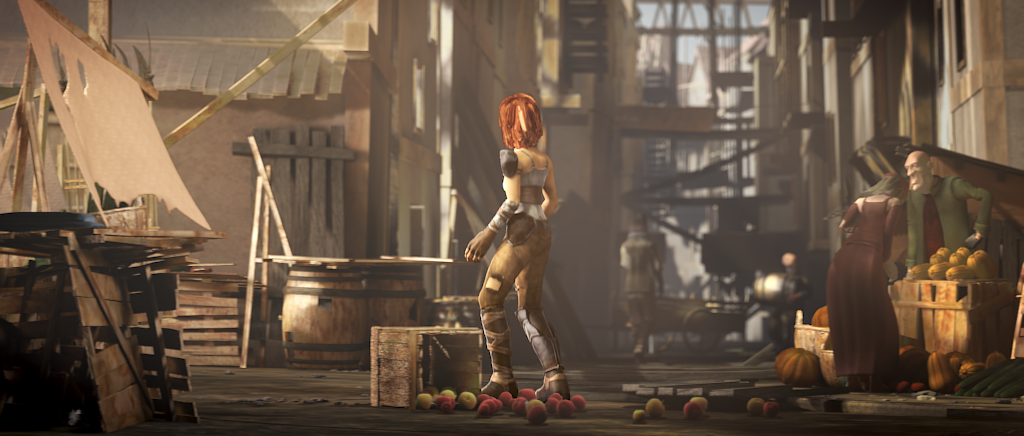}
        \caption{c) Input Frame}
    \end{subfigure}
    \hfill
    \begin{subfigure}[t]{0.24\textwidth}
        \includegraphics[width=\linewidth]{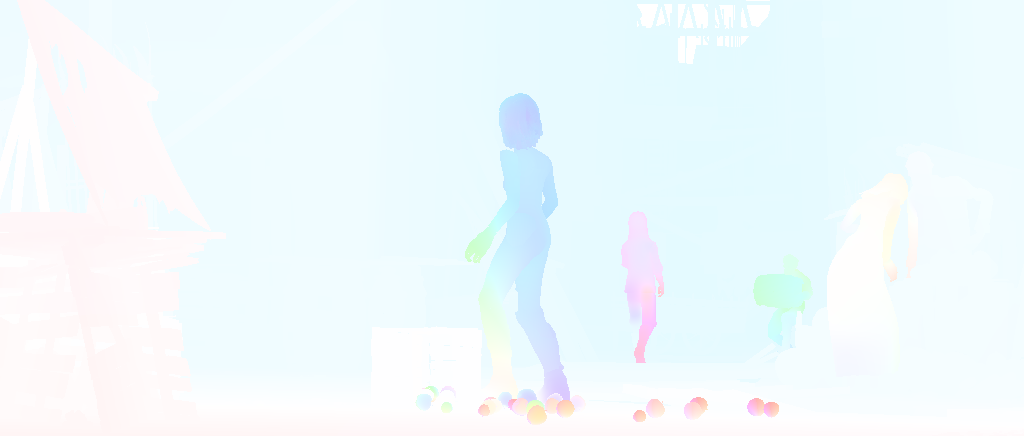}
        \caption{Ground Truth}
    \end{subfigure}
    \hfill
    \begin{subfigure}[t]{0.24\textwidth}
        \includegraphics[width=\linewidth]{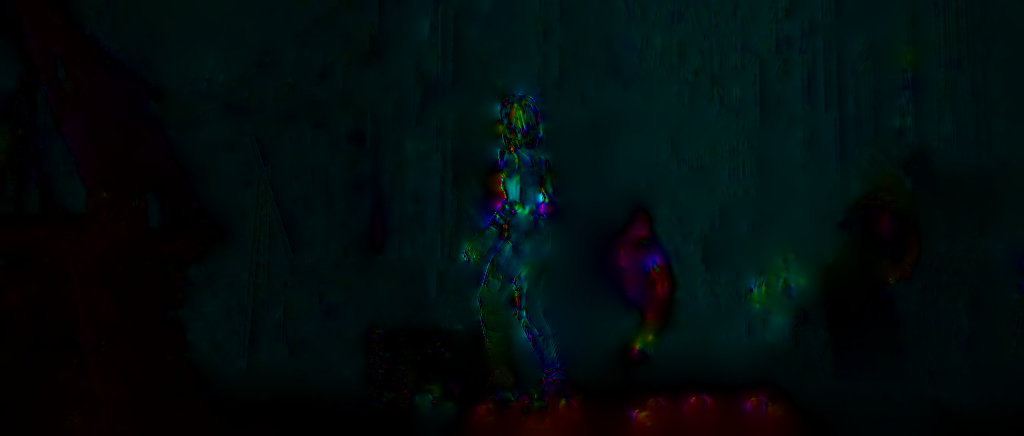}
        \caption{HS Flow}
    \end{subfigure}
    \hfill
    \begin{subfigure}[t]{0.24\textwidth}
        \includegraphics[width=\linewidth]{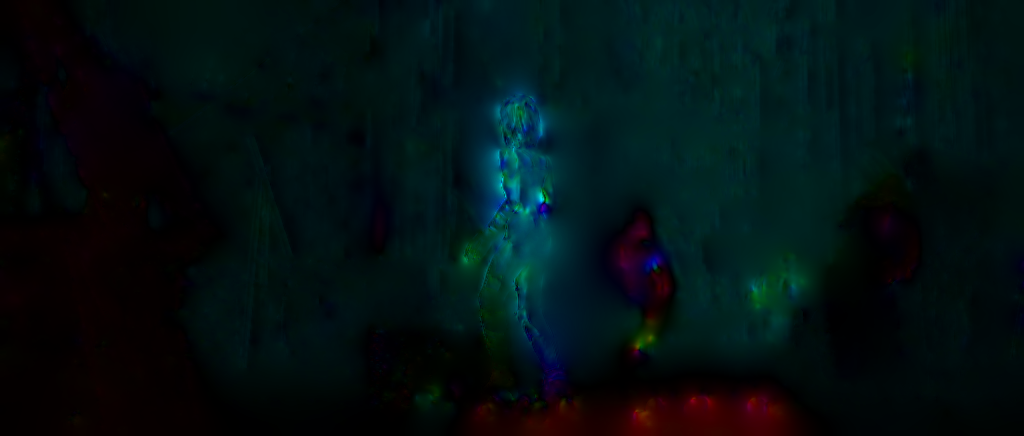}
        \caption{MR-HS Flow}
    \end{subfigure}
    \vspace{0.5em}
    \begin{subfigure}[t]{0.24\textwidth}
        \includegraphics[width=\linewidth]{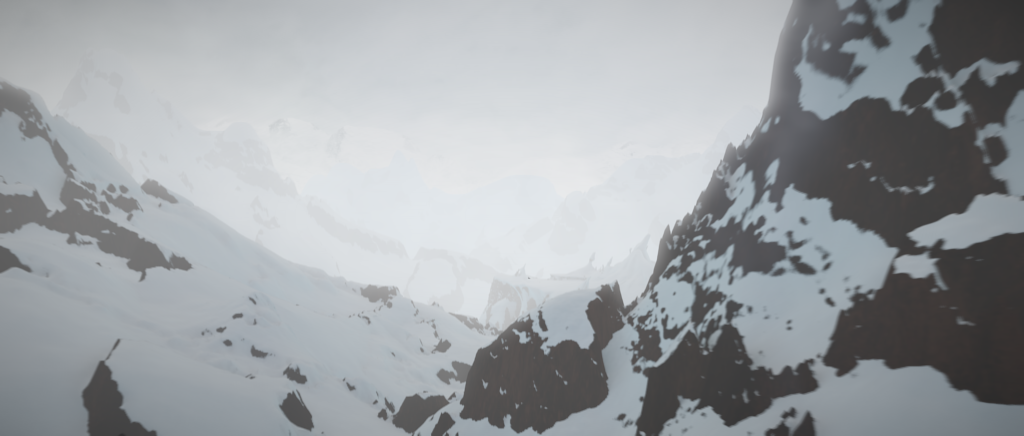}
        \caption{d) Input Frame}
    \end{subfigure}
    \hfill
    \begin{subfigure}[t]{0.24\textwidth}
        \includegraphics[width=\linewidth]{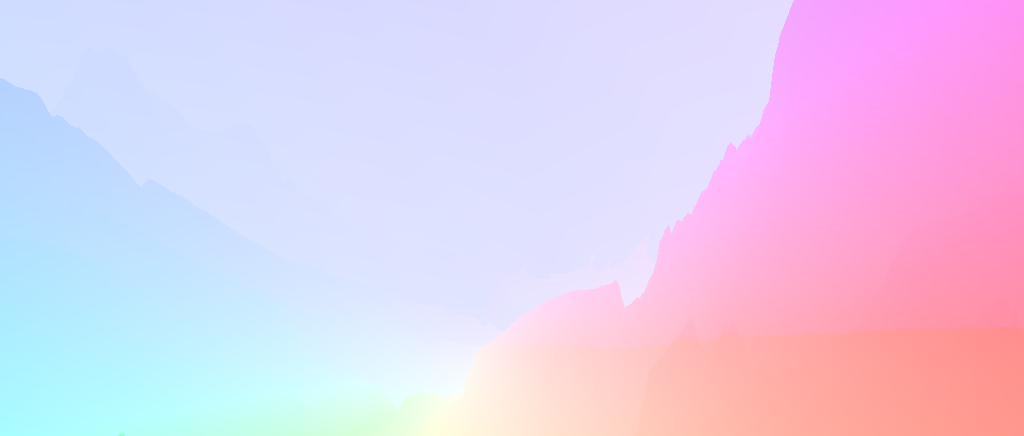}
        \caption{Ground Truth}
    \end{subfigure}
    \hfill
    \begin{subfigure}[t]{0.24\textwidth}
        \includegraphics[width=\linewidth]{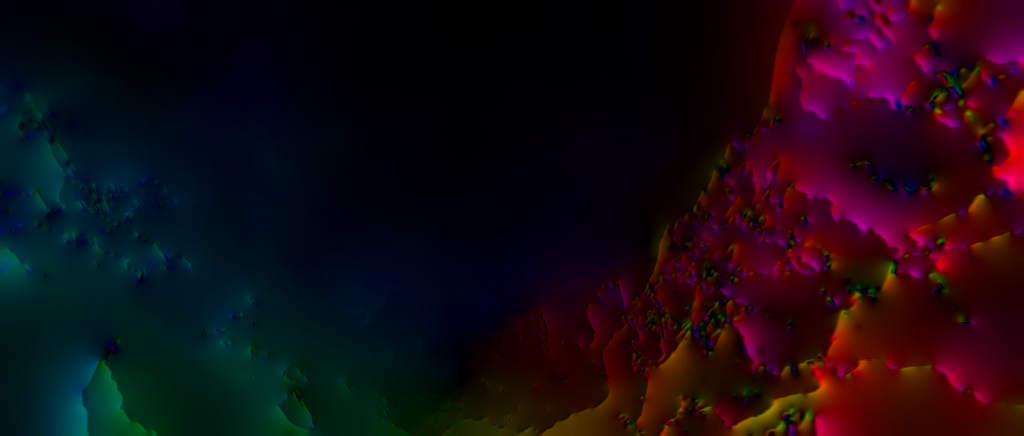}
        \caption{HS Flow}
    \end{subfigure}
    \hfill
    \begin{subfigure}[t]{0.24\textwidth}
        \includegraphics[width=\linewidth]{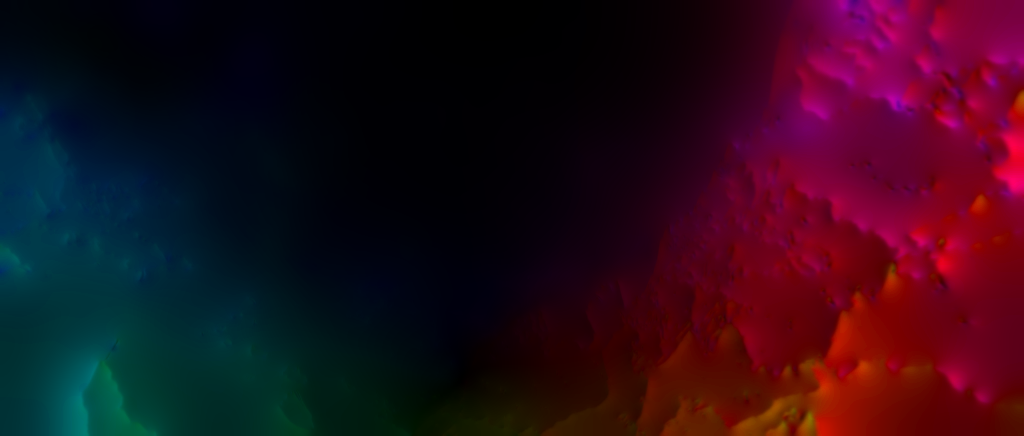}
        \caption{MR-HS Flow}
    \end{subfigure}
    \caption{Comparison of optical flow fields across four Sintel scenes (\textit{alley\_1 (frames 1, 2)}, \textit{bamboo\_2 (28, 29)}, \textit{market\_2 (41, 42)}, and \textit{mountain\_1(35, 36)}) using input frames, ground truth, Horn-Schunck (HS), and multi-resolution Horn-Schunck (MR-HS).}
    \label{fig:combined_flow_comparison}
\end{figure}

\newpage
The multiresolution Horn-Schunck (MR-HS) method demonstrates clear improvements over the original Horn-Schunck (HS) algorithm across all tested Sintel scenes. Notably, MR-HS consistently reduces both the Average Angular Error (AAE) and the End-Point Error (EPE), indicating more accurate motion estimation. These improvements are particularly significant in scenes with large displacements or complex textures, where the coarse-to-fine strategy of MR-HS and pyramid-based initialization avoids local minima and enhances convergence stability.

\section{Conclusion}

In this paper, we have presented an analysis of local and global methods for optical flow estimation, focusing on the Horn-Schunck algorithm and its multiresolution extension. The Lucas-Kanade method, a representative local technique, was discussed for its efficiency in sparse flow estimation, especially in textured regions. However, it was shown to struggle with large displacements and textureless areas, necessitating the incorporation of global methods.

The Horn-Schunck method, a global approach, was explored for its ability to enforce spatial coherence through smoothness priors. Despite its advantages, it is limited by issues near motion boundaries and large displacements. These challenges were addressed through the adoption of a multiresolution framework, which enhances the robustness and accuracy of the Horn-Schunck method by progressively refining estimates at multiple scales. The use of bilinear interpolation for prolongation between pyramid levels further improved the method's performance, especially in handling large motion and ensuring smooth transitions across scales.

Future work may include exploring other regularization techniques, such as total variation models, and further refining the multiresolution framework to improve its performance in real-world scenarios with complex motion. Additionally, combining the strengths of sparse and dense methods could lead to more efficient and accurate optical flow estimation techniques, especially for large-scale applications in computer vision and robotics.

Overall, this paper demonstrates the importance of combining local and global methods for optical flow estimation, emphasizing the need for multiresolution approaches to handle diverse motion patterns in dynamic environments.

\newpage
\section*{Acknowledgments}

I would like to express my sincere gratitude to my supervisor, Dr. El Mostafa Kalmoun, for his invaluable guidance, insightful feedback, and  support throughout this entire process. His expertise and encouragement have been instrumental to the success of this work, and it would not have been possible without his contributions.

\end{document}